%% file: lrec-coling2024-example.tex
\documentclass[10pt, a4paper]{article}

\usepackage{lrec-coling2024} 
\usepackage{subcaption}
\usepackage{longtable}
\usepackage{array}
\newcolumntype{P}[1]{>{\raggedright\arraybackslash}p{#1}}
\title{Benchmarking GPT-4 on Algorithmic
Problems:\\ A Systematic Evaluation of Prompting Strategies}

\name{Flavio Petruzzellis\textsuperscript{1}, Alberto Testolin\textsuperscript{1,2}, Alessandro Sperduti\textsuperscript{1}}

\address{\textsuperscript{1}Department of Mathematics, University of Padova, Padova, Italy \\ \textsuperscript{2}Department of General Psychology, University of Padova, Padova, Italy\\
flavio.petruzzellis@phd.unipd.it, alberto.testolin@unipd.it, alessandro.sperduti@unipd.it\\}

\abstract{
\input{0_abstract}
 \\ \newline  }

\begin{document}

\maketitleabstract

\section{Introduction}
\input{1_introduction}

\section{Related works}
\input{5_related}

\section{Reasoning Tasks}
\input{2_task}

\section{Models and Methods}
\input{3_methods}

\section{Results}
\input{4_results}

\section{Conclusion}
\input{6_conclusion}

\section{Ethics statement}
We do not recognize any ethical or legal issue regarding the data used in the present study, since we have analyzed and processed only synthetic, non-personal data.

While we acknowledge that an improper use of Large Language Models can have a major impact on society, for example potentially facilitating the spread of misinformation, we do not foresee any serious risk connected to the use of Large Language Models in the domain of algorithmic reasoning.

\section{Acknowledgements}
We are grateful to OpenAI for granting free research access to the GPT-4 and GPT-3.5 APIs.

\nocite{*}
\section{Bibliographical References}\label{sec:reference}

\bibliographystyle{lrec-coling2024-natbib}
\bibliography{bibliograpy}

\onecolumn
\newpage
\section{Appendix}
\input{7_appendix}

\end{document}

%% file: 0_abstract.tex
Large Language Models (LLMs) have revolutionized the field of Natural Language Processing thanks to their ability to reuse knowledge acquired on massive text corpora on a wide variety of downstream tasks, with minimal (if any) tuning steps.
At the same time, it has been repeatedly shown that LLMs lack systematic generalization, which allows to extrapolate the learned statistical regularities outside the training distribution.
In this work, we offer a systematic benchmarking of GPT-4, one of the most advanced LLMs available, on three algorithmic tasks characterized by the possibility to control the problem difficulty with two parameters.
We compare the performance of GPT-4 with that of its predecessor (GPT-3.5) and with a variant of the Transformer-Encoder architecture recently introduced to solve similar tasks, the Neural Data Router. 
We find that the deployment of advanced prompting techniques allows GPT-4 to reach superior accuracy on all tasks, demonstrating that state-of-the-art LLMs constitute a very strong baseline also in challenging tasks that require systematic generalization.

%% file: 1_introduction.tex
In very recent years, Large Language Models (LLMs) have become ubiquitous in natural language processing applications, thanks to their high degree of flexibility which allows to apply them on downstream tasks either directly or with limited amount of adaptation \citep{DBLP:conf/nips/BrownMRSKDNSSAA20}.
At the same time, since their first appearance it has been shown that LLMs struggle in reasoning tasks requiring several thinking steps to arrive to the final answer.
In this work, we focus on a specific subset of such tasks, namely algorithmic reasoning tasks, in which the problem samples can be automatically generated.
These problems can be solved with relatively simple algorithms, and their complexity can be fully characterized in terms of the parameters used in the procedure that generated the problem instances.
While LLMs are typically used to solve tasks that require processing unstructured natural language, studying their capacity to solve algorithmic reasoning problems can shed lights on their current limitations and therefore contribute to improve their design or prompting techniques.

Here, we consider three algorithmic tasks sharing the same general structure, but with different degrees of complexity.
All tasks require to simplify a  formula up to a minimal form. Formulas belong to three different domains: operations on list of integers (adapting the pre-existing ListOps dataset by \citealp{DBLP:conf/naacl/NangiaB18}), arithmetic, and algebra.
In all tasks the data distribution can be parameterized by the number of nested operations and the number of operands for each operation appearing in the formulas.

In order to benchmark the ability of GPT-4 to solve this sort of problems, we take into account seven different prompting techniques that have been recently proposed to solve reasoning tasks with LLMs. In the case of the two most advanced prompting methods, we further compare the performance of GPT-4 with that of its predecessor (GPT-3.5).
As an additional baseline, we also evaluate the performance of a recently proposed variant of the Transformer-Encoder architecture: the Neural Data Router \citep{DBLP:conf/iclr/CsordasIS22}, which has been specifically designed to solve reasoning problems consisting of formulas that require to be iteratively simplified.

We find that all models fail to solve the algorithmic problems we consider when they are generated using deeply nested formulas, with three or more operands for each operation.
We also observe that prompting techniques that lead the model to explicitly produce intermediate reasoning steps result in higher accuracy, especially in the case of arithmetic.
Furthermore, analyzing in detail the impact of each prompting method on the capacity of GPT-4 to generalize systematically, we find that the performance of the model improves mainly on formulas with low levels of complexity in terms of number of operands and nesting depth.

%% file: 5_related.tex
\textbf{Generalization in Transformers.} The characterization of generalization capabilities in Transformer-based architectures has been an active research area in the past few years.
One major research stream deals with out-of-distribution generalization, for example by investigating the capability of Transformers to achieve good performance on samples longer than those seen during training \citep{DBLP:conf/nips/AnilWALMRSGDN22,DBLP:journals/corr/abs-2306-15400}. Another major research topic is the study of compositional (also called systematic) generalization in Transformers \citep{DBLP:journals/jair/HupkesDMB20}, which investigates their capability to learn elementary solution procedures and compose them to solve more complex problems.

One element that has been shown critical to improve out-of-distribution generalization is the kind of Positional Encodings used in the Transformer.
In particular, label-based encodings \citep{DBLP:journals/corr/abs-2210-00400} extended out-of-distribution generalization in simple synthetic tasks, and similar work provided further experimental support for such method \citep{DBLP:conf/acl/RuossDGGCBLV23}.
A systematic empirical study comparing the length generalization performance of decoder-only Transformers using several different kinds of positional encodings has been recently presented by \citet{DBLP:journals/corr/abs-2305-19466}.
Another element which has been found to influence the capacity of Transformers to generalize on problem lengths is the compostion of the training set, which has been studied by \citet{DBLP:journals/corr/abs-2210-03275} on Sudoku puzzles, by \citet{DBLP:conf/acl/PatelBBG22} on the SCAN dataset, and by \citet{DBLP:journals/corr/abs-2306-15400} on arithmetic sum and multiplication.

Other work studied the impact of simple architectural choices \citep{DBLP:conf/emnlp/CsordasIS21,DBLP:conf/acl/OntanonAFC22}, including positional encodings, on the compositional generalization capability of Transformers. 
In \citep{DBLP:conf/emnlp/CsordasIS22} the authors present an extension of the CTL task designed to test models on unseen compositions of known functions.

\textbf{Prompting methods for systematic reasoning.}
A hot topic in Large Language Models is the design of effective prompting techniques.
\citep{DBLP:conf/nips/AnilWALMRSGDN22} offers a general comparison of different methods to improve out-of-distribution generalization in LLMs, including prompting techniques and fine-tuning.

Several approaches have explored the idea to let LLMs use the context -- including their own output -- to improve their reasoning and generalization capability.
A simple yet effective way to implement this idea is Chain-of-Thought prompting \citep{DBLP:conf/nips/Wei0SBIXCLZ22}, which has been further studies in scenarios with limited availability of prompting rationales \citep{DBLP:conf/nips/ZelikmanWMG22}.
In such case, the authors show that it is possible to boost performance by iteratively fine-tune the models on effective rationales produced by the models' themselves.

Other work has further built on the idea of leveraging the models outputs to solve reasoning problems: \citet{DBLP:journals/corr/abs-2112-00114} introduced the idea of a scratchpad, i.e. an arbitrary sequence of intermediate tokens which usually follows a specific formatting and that can be used by the model to more explicitly elaborate the problem before producing the answer.
More recently, \citet{DBLP:journals/corr/abs-2211-09066} introduced Algorithmic Prompting, a method by which highly-detailed solution procedures for arithmetic problems are provided to the model in a few-shot regimen.
A more general case of scratchpad has been presented by \citet{DBLP:journals/corr/abs-2305-00833}, allowing LLMs to produce intermediate outputs also while receiving the input, mimicking a note-taking behavior. 
Finally, \citet{DBLP:conf/iclr/ZhouSHWS0SCBLC23} showed that reasoning in LLMs can be improved by iteratively prompting the models to first decompose a problem into sub-problems, solve each sub-problem separately, and then gradually compose the final solution.

%% file: 2_task.tex
\begin{table*}
    \centering
    \input{tables/inputs_examples}
    \caption{Examples of inputs of the arithmetic task for the nine data splits considered. The complexity of formulas can be manipulated by increasing the number of operands for each operation and by increasing the depth of nesting points, which can occur several times in the same formula.}
    \label{tab:inputs_examples}
\end{table*}

Reasoning tasks such as solving school-level math word problems, commonsense reasoning or algorithmic reasoning have always been one of the hardest class of tasks for Large Language Models.
Such problems necessitate several `thinking steps' to arrive at the final solution, which usually needs to be derived from intermediate calculations.

The machine learning community has proposed a variety of tasks specifically designed to study the ability of neural architectures to generalize in a systematic and compositional way \citep{DBLP:conf/icml/LakeB18, DBLP:journals/corr/abs-1802-06467, DBLP:conf/emnlp/KimL20}.

Inspired by these tasks, we consider the general problem of iteratively simplifying a formula.
We consider a problem framework in which it is possible to generate synthetic formulas with different levels of complexity, which can be characterized by two parameters: the maximum nesting depth of any operation in a formula (Nesting), and the maximum number of operands involved in each operation (Operands).
Given its generality, this problem framework can be applied to different domains where nested  formulas with an arbitrary number of operands can be defined.
We thus define three tasks with different levels of complexity: operations on lists of integers, arithmetical operations and algebraic operations.

By varying the values of the Nesting and Operands parameters, we could define an arbitrary number of \textit{data splits} for each task, each featuring a different level of difficulty.
In our experiments, we consider the nine data splits which result from taking the values in the Cartesian product of the sets $N=\{2, 3, 4\}$ and $O=\{2, 3, 4\}$, representing values of the Nesting and Operands parameters, respectively.
In Table \ref{tab:inputs_examples} we report, as a reference, examples of formulas from the arithmetic task which have been drawn from the nine data splits used throughout the experiments.
As the examples show, a formula can be nested in several points of its structure, meaning that there can be multiple non-nested operations which can be simplified at any time.

\subsection{ListOps}
The ListOps dataset \citep{DBLP:conf/naacl/NangiaB18} was proposed as a simple benchmark to assess the capacity of neural networks to evaluate nested expressions having parse trees of different depths.
The original task included four simple operations (minimum, maximum, sum modulo 10 and median), which are applied to lists of single-digit integers.
The final solution of a problem is always a single-digit integer.
For example, one instance of the problem might be \texttt{[MIN[MAX24567][SM10293]213]} having as a unique solution the number \texttt{1}.
In order to adapt this dataset to the problem framework described above, we extend the problem definition by making it possible to specify also the number of operands appearing in the operations, other than the operations' nesting depth.
At the same time, we slightly simplified the problem by considering only three operations: minimum, maximum and sum modulo 10.

\subsection{Arithmetic}
The second type of task consists in finding the final value of an arithmetic expression where the operators involved are sum, subtraction and multiplication, and operands are integers in the interval $(-100, 100)$.
To keep the problem complexity at a reasonable level and focus on the capacity of the model to solve a complex problem by applying simple solution steps recursively, each intermediate value obtained in the solution process was taken modulo 100.

\subsection{Algebra}
We finally adopted the general problem structure to the domain of symbolic mathematics, considering a subset of algebraic expressions in which all formulas can be reduced to a minimal form, i.e. either a single number, a monomial, or a binomial.
We automatically generate algebraic expressions consisting of sums and subtractions between monomials.
Each monomial can contain up to four variables and has a numerical coefficient in the range $(-100, 100)$.
For example, the formula $(((30xy+33xy)+(-80xy+62xy))-62xy)$ is sampled from the data split parameterized by $(N=3, O=2)$ and its simplified form is $-17xy$.

%% file: tables/inputs_examples.tex
\begin{tabular}{p{1.5cm}p{2.5cm}p{2.7cm}p{7.3cm}}
 & 2 Operands & 3 Operands & 4 Operands \\
\hline
Nesting 2 & ((-21+47)*\newline(38*-69)) & (-73-(33*54)+55) & ((-28+32)-(28-11+65)+(13+53)-(-15*20)) \\
\hline
Nesting 3 & (57*((5+1)+\newline(-79+60))) & (((35-2+12)-94+(62*-30))+\newline((-97*-75)-\newline(-10*-53)+9)-74) & (((-6-41-91-80)-(-31*-22)-(-54*84)-(0+77))+\newline((-77-27)-77-86-96)+(91+20+(-3+3-30)+\newline(-41-65+6+89))-((-83-23+50)+34-(-93+4-15-8)-(35*-26))) \\
\hline
Nesting 4 & (-35*(((27*53)+\newline(-43*-51))+\newline((-19*81)+\newline(42*66)))) & ((((-86+25)\newline-(-87+76-8)-\newline(17-93+19))+\newline((-22-79-17)+\newline72+4)+(-80-(-96*\newline-15)-64))-32-36) & (((-66+(-52*51)+43-(-62+69+81+38))-((97*83)+86-41-85)-((91+8+89)+\newline(-15+33+99)+12+(-6-53+18))-\newline(-48-(64+77+36+69)+(-56+12-80)-27))+(((-74+7)+(49+96-4)-(20-1)-(72-5-78))-(16+69+(59-61+80+9)+(78+60+3))-(46+(19+10-48+14)+(61*-4)+(0+86+40-4))+\newline(-53-79+(31*-94)-68))+(-16+81+71+\newline(-55-41+(-12*-73)-32))+(84-74+((13-27+17-90)-(15+75+93)+(54+37-62)+(71-23+46-4))-((61+14)-(-32-87)+(68-22-25)-(14*-7)))) \\
\end{tabular}

%% file: 3_methods.tex
In the following sections, we briefly describe the prompting techniques we used to probe GPT models, as well as the structure of the Neural Data Router \citep{DBLP:conf/iclr/CsordasIS22} and the minimal modifications we have made to adapt the architecture to the problems at hand.

\subsection{Prompting techniques}

In the following, we present the prompting techniques involved in this study.
Examples of each prompting technique on the three tasks are reported in paragraph \ref{subpar:app2} in the Appendix.

\subsubsection{Zero-shot}
This prompting technique simply consists in giving the problem description as input to the model and directly asking for the result.
The question is formatted in such a way that the output of the model will be constrained to the desired format and can thus be easily parsed.

\subsubsection{Role assignment}
We also experiment with a variant of the Zero-shot prompting technique in which we assign a role to the agent.
Following recent findings \citep{DBLP:journals/corr/abs-2308-07702} which suggest that specifying the agent's field of expertise could improve the accuracy of its answers, we input the sentence ``You are a brilliant mathematician'' before giving the model the actual problem description.

\subsubsection{Few-shot}
One of the abilities that had a big impact on the popularization of LLMs is their capacity to learn from examples at inference time, a technique called `few-shot' or `in-context' learning \citep{DBLP:conf/nips/BrownMRSKDNSSAA20}.
In this case, we prompt the models by providing a list of three examples of solved formulas before asking to solve the actual problem. 
The formulas are sampled from three data splits described by the following values of the Nesting and Operands parameters: $(N=1, O=2)$, $(N=2, O=2)$ and $(N=2, O=3)$.
In the examples, the formulas are solved directly, i.e. without showing intermediate solution steps to the model.

\subsubsection{Chain-of-Thought}
One of the most general and effective techniques that have been recently proposed to elicit reasoning in Large Language Models is Chain-of-Thought (CoT) prompting \citep{DBLP:conf/nips/Wei0SBIXCLZ22}.
When prompted following this method, the model receives a set of examples showcasing the solution of a given reasoning problem, where each example explicitly includes the intermediate solution steps required to get to the final answer.

In our case, we experiment with two different kinds of CoT prompting: in the first one, named `Symbolic Chain-of-Thought', we provide the solution examples to the model exclusively in a symbolic form, that is, as a chain of equalities.
In the second case, named `Verbal Chain-of-Thought', each intermediate step is also described with English text, suggesting the model a motivation for taking that simplification step and encouraging it to mimic the same verbalization behavior when producing the answer to the actual problem. 

\subsubsection{Zero-shot Chain-of-Thought}
Zero-shot Chain-of-Thought prompting \citep{DBLP:conf/nips/KojimaGRMI22} is a technique which has been proposed to obtain similar results as the ones obtained with Chain-of-Thought prompting, without the need to carefully engineer prompts with examples demonstrating the solution steps.
The model is prompted directly with the problem it needs to solve, as well as with the first words of the answer: ``Let's think step-by-step''.
The output generated by the model is then collected and used to prompt the model a second time to get the final answer in the correct format, now eliciting the output with  the usual formula: ``So, the final answer is:''.

\subsubsection{Self-consistency}
Reasoning problems are different from other problems that can be tackled with generative models, in that they always have a unique solution (at least semantically, i.e. not taking into account the different ways in which a solution can be written, for example in the case of algebraic expressions).
Nevertheless, there could be multiple reasoning paths that lead to the correct solution, varying not only formally, but also substantially -- as in, for example, theorem proving.
Starting from this premise, \citealp{DBLP:conf/iclr/0002WSLCNCZ23} advocate for \textit{self-consistency} in Large Language Models' outputs when solving reasoning tasks.
The basic idea is that the performance of the model might improve if we prompt the model several times, and consider the answer that was generated more frequently and therefore, one might say, with more confidence.

We apply the self-consistency prompting method in combination with Zero-shot CoT prompting.
To limit the consumption of credits to query the OpenAI API, we prompted the model only 5 times for each input, rather than 40 times as done in the original work.
We note, therefore, that the models performance might further improve raising the number of prompts per input, and thus the confidence in the selected answer.
However, even with such a small number of outputs, we can already observe the effectiveness of this prompting technique.

\subsection{Neural Data Router}
The Neural Data Router \citep{DBLP:conf/iclr/CsordasIS22} has recently been proposed as a modification of the Encoder module in the Transformer architecture \citep{DBLP:conf/nips/VaswaniSPUJGKP17} with the specific goal of solving problems where applying the same resolution step iteratively (such as solving a sub-expression in a complex formula) can lead to the final solution.
The modifications introduced in this model are conceived to allow systematic generalization on such problems.
The first one, named `copy gate', is a mechanism that allows the model to entirely skip the computation in the self-attention and feed-forward blocks of an Encoder layer, and directly transmit the input to the following layer.
The second mechanism, called `geometric attention', is a modification of self-attention designed to facilitate the focus on the closest match of any token, thus allowing to consider only a narrow region of the input sequence which should ideally correspond to the part of formula to be solved.

The model was originally tested on three tasks: ListOps \citep{DBLP:conf/naacl/NangiaB18}, Compositional Table Lookup (CTL) \citep{DBLP:journals/corr/abs-1802-06467}, and arithmetic formulas with single-digit operands and intermediate values taken modulo 10.

In the original work, the final result could be collected in the first or last position of the encoded sequence, since these problems always have single-digit integers as targets.
This no longer holds in our case, as the algorithmic problems we consider can have a solution containing more than one token. We thus adapted the model by modifying the mechanism of collection of the final result, considering a window at the beginning of the final sequence produced by the encoder which is as large as the expected target.

For all tasks, we include in the training set examples drawn from the data defined by the parameters values $(N=1, O=1)$, $(N=1, O=2)$, $(N=2, O=2)$ and $(N=2, O=3)$.
The training sets are balanced across the four different data splits and include $400,000$ samples for all tasks.

We select the model's hyperparameters reproducing the search procedure described in the original work.
We define two validation sets: an in-distribution validation set which mirrors the composition of the training set, and an out-of-distribution  validation set which includes samples from more difficult splits, represented by the parameters \mbox{$(N=2,O=4)$}, \mbox{$(N=3,O=2)$}, \mbox{$(N=3,O=3)$}, \mbox{$(N=3,O=4)$}, \mbox{$(N=4,O=2)$}, \mbox{$(N=4,O=3)$} and \mbox{$(N=4,O=4)$}.
Both validation sets are balanced across data splits and include $1,000$ samples per data split, following the same model selection protocol used in the original work.
We select the best performing model on the out-of-distribution validation set applying the Bayesian hyperparameter search tool provided by the Weights and Biases MLOps platform\footnote{\url{https://wandb.ai}} to optimize the following hyperparameters: learning rate, number of encoder layers, dimensionality of the hidden state, number of heads, weight decay, dropout, attention dropout and size of the hidden feed-forward layer.
The parameter ranges used in the hyperparameters search procedure are the same reported in the original work.

It should be noted that the scenarios in which we probed this model can be considered more challenging than the ones considered in the original work.
Indeed, we used a smaller training set for the ListOps task (the original training set included formulas with nesting depth up to 5), and we considered more complex arithmetic expressions, as well as algebraic expressions on which the model was never tested before.

%% file: 4_results.tex
\begin{table}
    \centering
    \input{tables/summary_table}
    \caption{Average performance of all models and prompting methods on both in-distribution and out-of-distribution test splits, measured in terms of percentage accuracy. ``Zero-shot role'' refers to the Zero-shot prompting method where the agent  was assigned a role. The best performance for each task is highlighted in bold.}
    \label{tab:perf_aggr}
\end{table}

In this section, we present the performance achieved by all models on the three algorithmic tasks. We also analyze the factors that might determine their success or failure, taking into account the characteristics of each task, the different level of complexity of the nine data splits, and the features of the prompts we used for GPT models.

In the case of GPT-3.5, we tested the model version  called     \texttt{gpt-3.5-turbo} available via the OpenAI API at the time of writing.

We test all models on test sets composed of 100 samples for each data split we consider, thus summing up to 900 samples for each task.
The performance of all models is measured in terms of response accuracy: the output produced by the model is considered correct only if it exactly matches the target.
The only exception is in the Algebra task, for which we relax this requirement and consider as correct also outputs that are semantically equivalent to the target, but written in a different form.
To carry out such semantic comparison between output and target, we employ the SymPy Python library\footnote{\url{https://www.sympy.org}} for symbolic mathematics.

A summary of the performance of all models and prompting methods on the test splits is reported in Table \ref{tab:perf_aggr}.
We can generally observe that none of the considered models or prompting methods were able to perfectly solve the problems at hand, demonstrating that systematic generalization is challenging for this sort of algorithmic problems.

\begin{figure}
    \centering
    \includegraphics[width=\columnwidth]{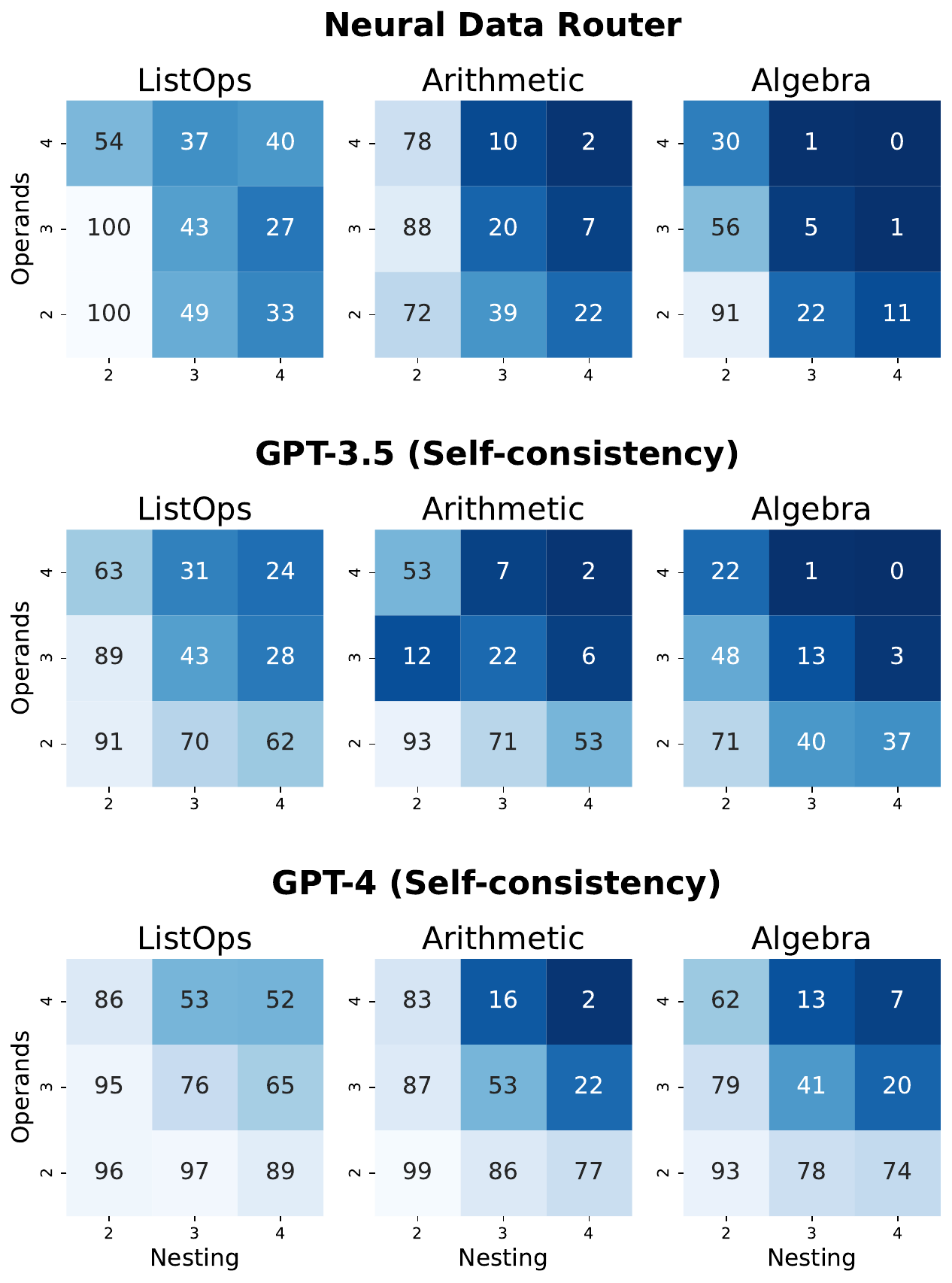}
    \caption{Performance of the Neural Data Router, GPT-3.5 and GPT-4 using Self-consistency prompting on the test splits. Values represent output accuracy in percentage: the performance of all models and prompting methods clearly decreases on data splits of higher complexity.}
    \label{fig:models-comparison}
\end{figure}

Figure \ref{fig:models-comparison} shows the performance of the Neural Data Router, GPT-3.5 and GPT-4 using Self-consistency prompting in a blown-out format, which allows to inspect the performance of all models on each data split.
We observe that all models achieve higher performance on simpler data splits, and that performance degrades more or less smoothly as the problem complexity (i.e., depth of formula and number of operators) increases.

In the following, we examine the performance of the Neural Data Router comparing it to that of the two versions of GPT.
Then, we consider the effectiveness of the seven prompting methods on each task, as well as their impact on the capacity of LLMs to apply solution strategies that generalize systematically to complex problems.

\subsection{Neural Data Router}

\begin{figure*}
    \centering
    \includegraphics[width=\textwidth]{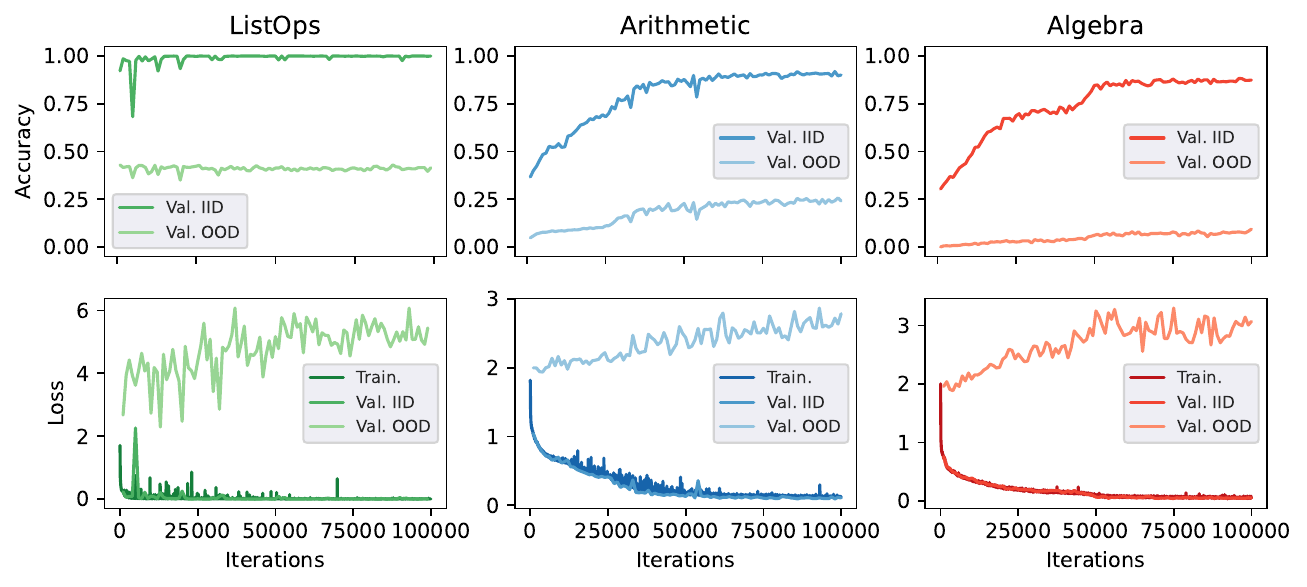}
    \caption{Accuracy and loss of the Neural Data Router during training on the three algorithmic tasks. Val. IID and Val. OOD refer to in-distribution and out-of-distribution validation sets, respectively. The model overfits the in-distribution split on all tasks, failing to generalize to more difficult samples.}
    \label{fig:ndr_training}
\end{figure*}

Figure \ref{fig:ndr_training} shows how accuracy and loss of the best Neural Data Router configuration resulting from the hyperparameters search evolve during training.
It is interesting to note that while the performance on the in-distribution validation set grows steadily, reaching more than $85\%$ accuracy on all tasks, the performance on the out-of-distribution validation set grows much more slowly, becoming almost constant when learning starts to converge on the training set.
The wide gap between the loss curves on the two validation sets shows that the model overfits the in-distribution split on all tasks, failing to generalize to more difficult samples.

Considering the detailed performance of the model represented in Figure \ref{fig:models-comparison}, we observe that, in the case of Arithmetic and Algebra, the model generalizes better on formulas with more operands than seen during training (data split \mbox{$(N=2, O=4)$)}, rather than in the case of more deeply nested formulas (data splits \mbox{$(N=3, O=2)$} and \mbox{$(N=4, O=2)$}).
Since the training set includes examples of a `base case' and an `induction step' for both parameters (respectively, \mbox{$(N=1, O=2)$} and \mbox{$(N=2, O=2)$} for Nesting, \mbox{$(N=2, O=2)$} and \mbox{$(N=2, O=3)$} for Operands), this could indicate that it is easier to learn a generalization step when the complexity of a formula increases in terms of number of operands, rather than in terms of nesting depth.

Table \ref{tab:perf_aggr} shows the overall performance of the model, including both in-distribution and out-of-distribution test splits.
While the model does not achieve the best performance overall, it is relevant to observe that its average performance is competitive with that of GPT-3.5 on all tasks. Further comparing the performance of the two models on each data split reported in Figure \ref{fig:models-comparison}, we observe that the Neural Data Router achieves a better performance on harder data splits on the ListOps task.
Therefore, while the overall performance of the model on this task is lower, this result suggests that in simple scenarios the mechanisms built in the NDR allow to learn a solution process that can generalize better than a large, general-purpose model such as GPT-3.5.

\subsection{Prompting methods and tasks}

As reported in Table \ref{tab:perf_aggr}, on all tasks the best performance was achieved by GPT-4 using the Self-consistency prompting method.
More generally, prompting techniques that require (or encourage) GPT-4 to reason explicitly were more effective: in all cases, the best performance obtained by prompting methods that ask the model to directly provide the answer (either with or without solution examples) is significantly lower than that achieved by Self-consistency prompting.
At the same time, it is interesting to note that by just contextualizing the role of the agent (Zero-shot role prompting) we can improve the model's accuracy in all tasks, compared to Zero-shot prompting.

Comparing Chain-of-Though prompts with symbolic solution steps and verbalized solutions steps, we found that asking the model to spell-out the reasoning process significantly improved the performance on the ListOps and Arithmetic tasks, while it had limited impact on the Algebra task.
Furthermore, we can see that the task for which CoT prompting is more useful is clearly Arithmetic, with a gain in accuracy up to $+40\%$ compared to Few-shot prompting.
While it is hard to single out a single reason why a given prompting method could be more effective in one particular task, we conjecture that Arithmetic could particularly benefit from the explicit formulation of solution steps due to the considerable difficulty of computing the product between two-digits integers, which for a Large Language Model is probably the hardest elementary operation present in our datasets.

\begin{figure*}
    \centering
    \includegraphics[width=\textwidth]{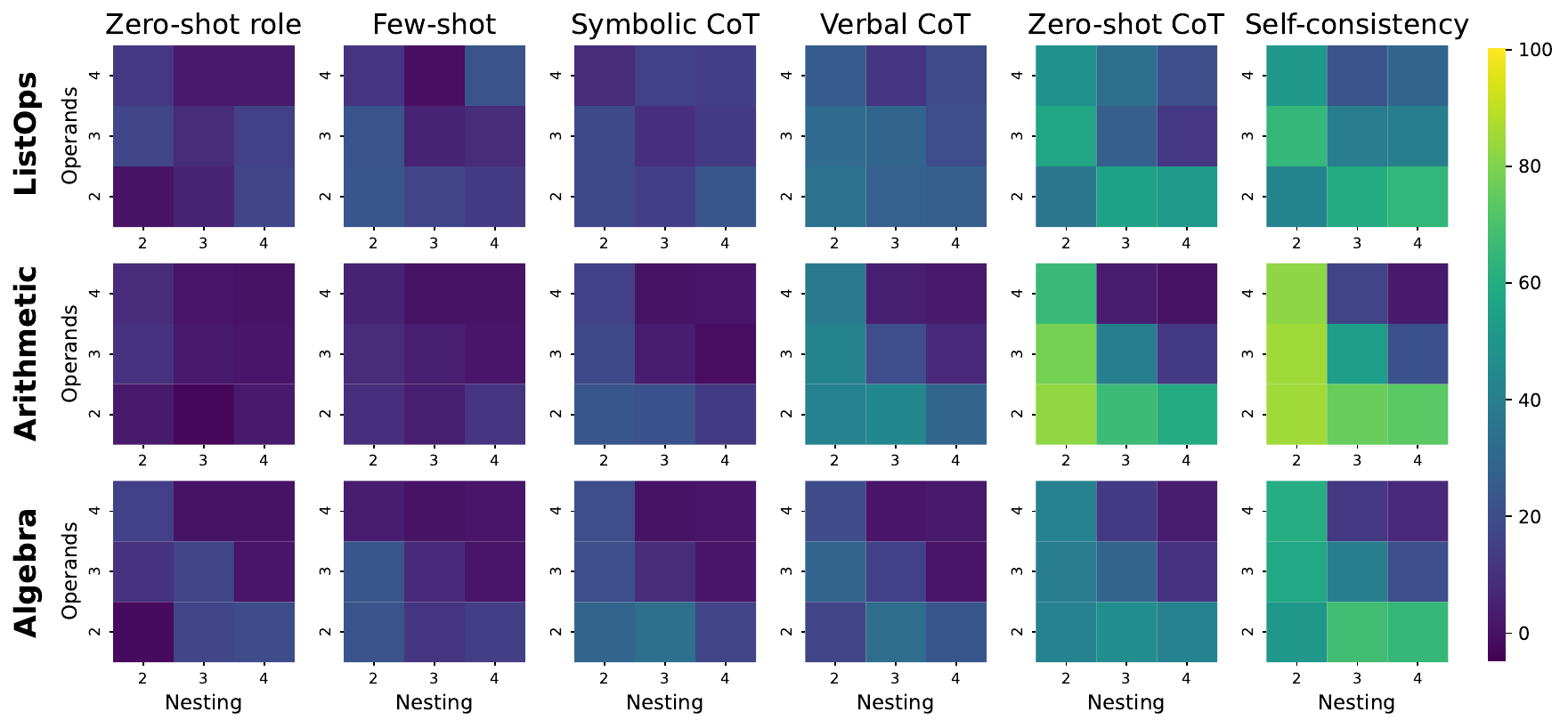}
    \caption{Performance gain measured as percentage accuracy resulting from each prompting method on GPT-4 compared to Zero-shot baseline. The accuracy gains from the best prompting methods are concentrated in simpler data splits, especially on Arithmetic.}
    \label{fig:gpt4-summary}
\end{figure*}

\subsection{Prompting methods and generalization}

Taking into account the performance of GPT-4 on the nine different data splits considered for each task, we can make some considerations on the capacity of different prompting methods to enable systematic generalization.

We show in Figure \ref{fig:gpt4-summary} the performance gain from each prompting method compared to the Zero-shot baseline, measured in terms of difference in percentage accuracy. We observe that, in general, the performance gains resulting from the prompting techniques that lead the model to make explicit reasoning steps are concentrated in the data splits with low levels of complexity.
The performance gain on data splits parameterized with $(N=3, O=4)$, $(N=4, O=3)$ and $(N=4, O=4)$ is generally less evident.

As previously noted, this general tendency is more evident on Arithmetic, where the performance gains deriving from prompting methods that produce explicit reasoning are greater overall.
On the other hand, the performance gain from such prompting methods is more evenly spread across data splits in the case of ListOps, for which the GPT-4 performance on simple splits is already quite high even using Zero-shot prompting, as shown in Table \ref{tab:perf_aggr} (see also paragraph \ref{subpar:app1} in the Appendix).

Overall, this suggests that the prompting methods we considered gradually improve the performance of GPT-4 on the problems at hand by increasing the model's effectiveness on simple problem instances, but they do not trigger the emergence of a solution mechanism that can lead to systematic generalization.

%% file: tables/summary_table.tex
\addtolength{\tabcolsep}{-0.4em}
\begin{tabular}{lrrr}
 & ListOps & Arithmetic & Algebra \\
\hline
GPT-3.5 & & & \\
\quad Zero-shot CoT & 0.44 & 0.32 & 0.19 \\
\quad Self consistency & 0.56 & 0.35 & 0.26 \\
\hline
GPT-4 & & & \\
\quad Zero-shot & 0.33 & 0.04 & 0.10 \\
\quad Zero-shot role & 0.42 & 0.06 & 0.18 \\
\quad Few-shot & 0.46 & 0.08 & 0.19 \\
\quad Symbolic CoT & 0.48 & 0.14 & 0.24 \\
\quad Verbal CoT & 0.58 & 0.29 & 0.25 \\
\quad Zero-shot CoT & 0.71 & 0.49 & 0.39 \\
\quad Self-consistency & \textbf{0.79} & \textbf{0.58} & \textbf{0.52} \\
\hline
NDR & 0.54 & 0.38 & 0.24 \\
\end{tabular}

%% file: 6_conclusion.tex
In this work, we investigated the effectiveness of a variety of LLMs prompting methods on three algorithmic tasks designed to analyse how systematic reasoning capabilities might emerge in relation to an increase in problem complexity.
We also compared the performance of state-of-the-art LLMs with that of a much simpler neural architecture, the Neural Data Router, specifically designed to solve algorithmic problems characterized by a recursive structure.

We found that none of the models and prompting methods could exhibit proper systematic generalization capacities.
Although the performance of the ad-hoc Neural Data Router model was competitive with that of a general purpose language model such as GPT-3.5, it turns out that more advanced language models like GPT-4 currently represent the state-of-the-art for solving this type of reasoning tasks, even on relatively complex problem instances.
By comparing the performance of the different prompting methods, we found that explicitly producing reasoning steps in a verbal form can generally improve model performance on all tasks. 
However, our experiments also highlight the limitations these prompting methods, showing that their effectiveness could be limited for complex symbolic reasoning problems.

While synthetic algorithmic problems are quite far from the typical scenario of application of Large Language Models, they are useful for precisely characterizing the limitations of such models, as done in the present work. Future work could extend our analysis by expanding the set of prompting methods considered, for example by including Least-to-Most prompting \citep{DBLP:conf/iclr/ZhouSHWS0SCBLC23}, which could in principle be applicable to problems requiring to solve nested  formulas.
Other nuances of prompts themselves could also be taken into account, such as the phrasing used when contextualizing the agent role, the number of exemplars provided in the few-shot learning regimen, or the way rationales for Chain-of-Thought prompting are constructed.

Researchers working on the design of prompting techniques for Large Language Models could propose novel methods to improve the capacity of the models to apply solution strategies that generalize well on complex problem instances.
More precisely, future research could be dedicated to understand the extent to which current limitations depend on the ability to effectively retrieve and manipulate information in a very large context, a scenario which could become more and more common in future dialogue-based real-world applications of LLMs.

Another promising venue for future research would be to design ad-hoc Transformer variants that might promote the emergence of systematic generalization on algorithmic problems, taking into account both the role of architectural mechanisms and the composition of the training distribution on the models' capacity to generalize.
~
~
~

%% file: 7_appendix.tex
\subsection{Performance of other models and prompting methods}
\label{subpar:app1}
\begin{figure}[h]
    \centering
    \begin{subfigure}[b]{0.48\textwidth}
        \centering
        \includegraphics[width=\textwidth]{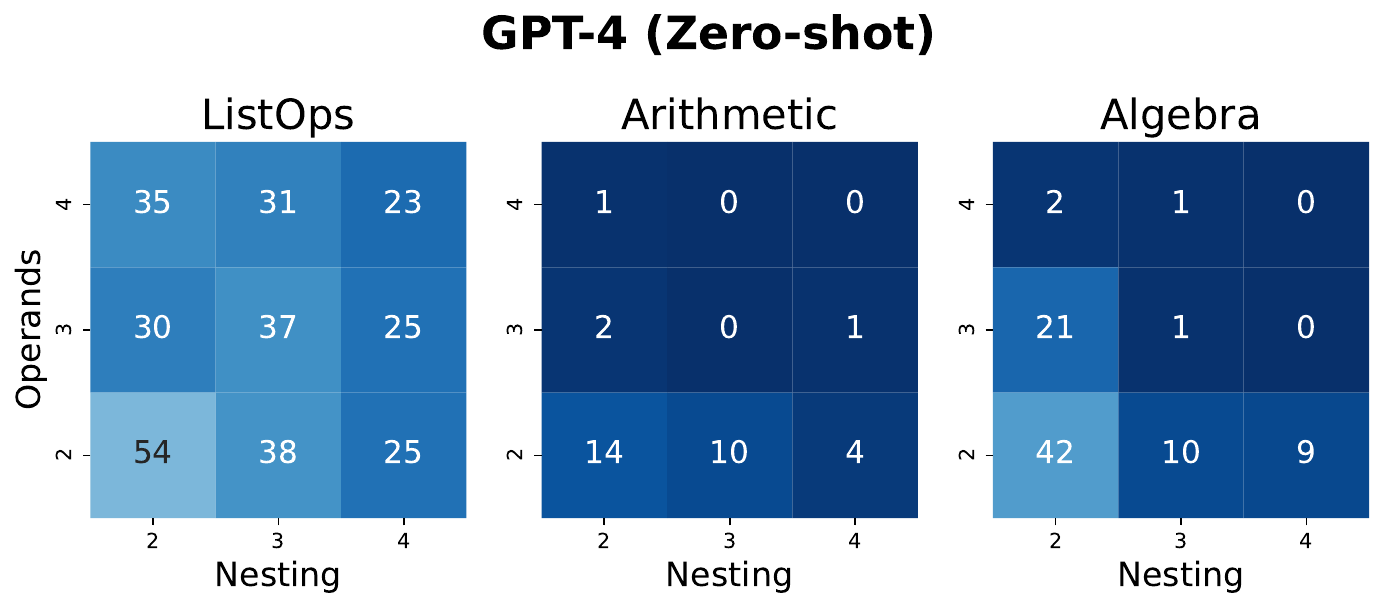}
    \end{subfigure}
    \hfill
    \begin{subfigure}[b]{0.48\textwidth}
        \centering
        \includegraphics[width=\textwidth]{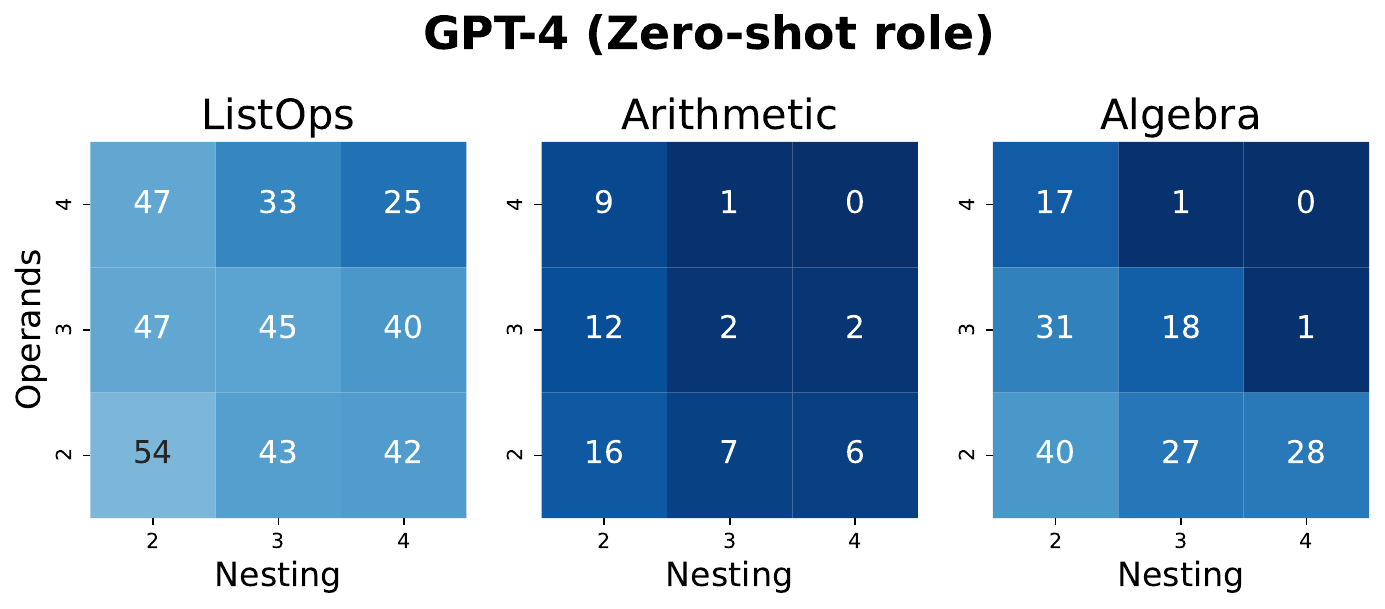}
    \end{subfigure}
    \begin{subfigure}[b]{0.48\textwidth}
        \centering
        \includegraphics[width=\textwidth]{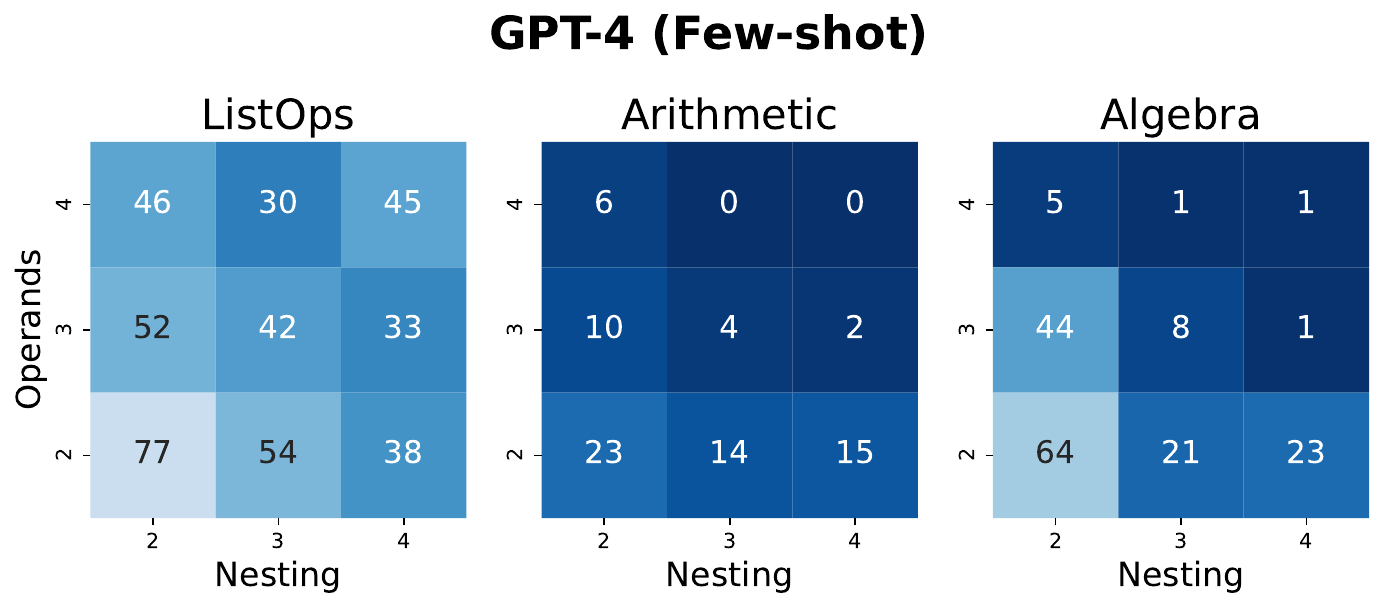}
    \end{subfigure}
    \hfill
    \begin{subfigure}[b]{0.48\textwidth}
        \centering
        \includegraphics[width=\textwidth]{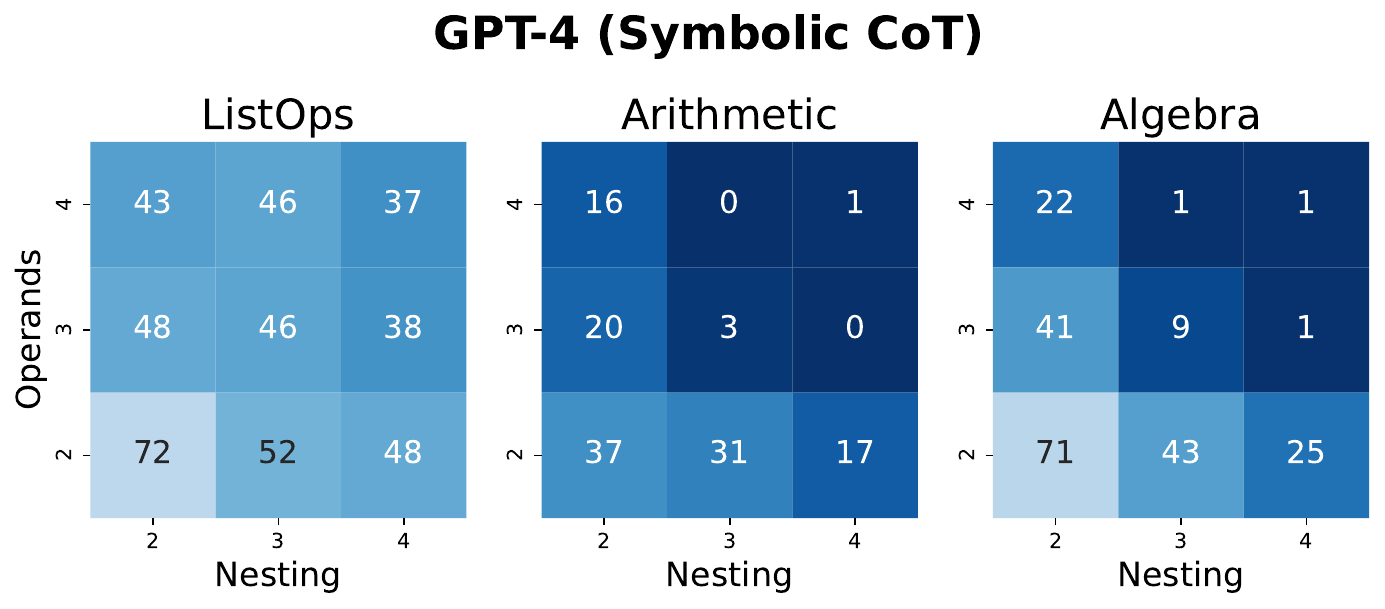}
    \end{subfigure}
    
    \begin{subfigure}[b]{0.48\textwidth}
        \centering
        \includegraphics[width=\textwidth]{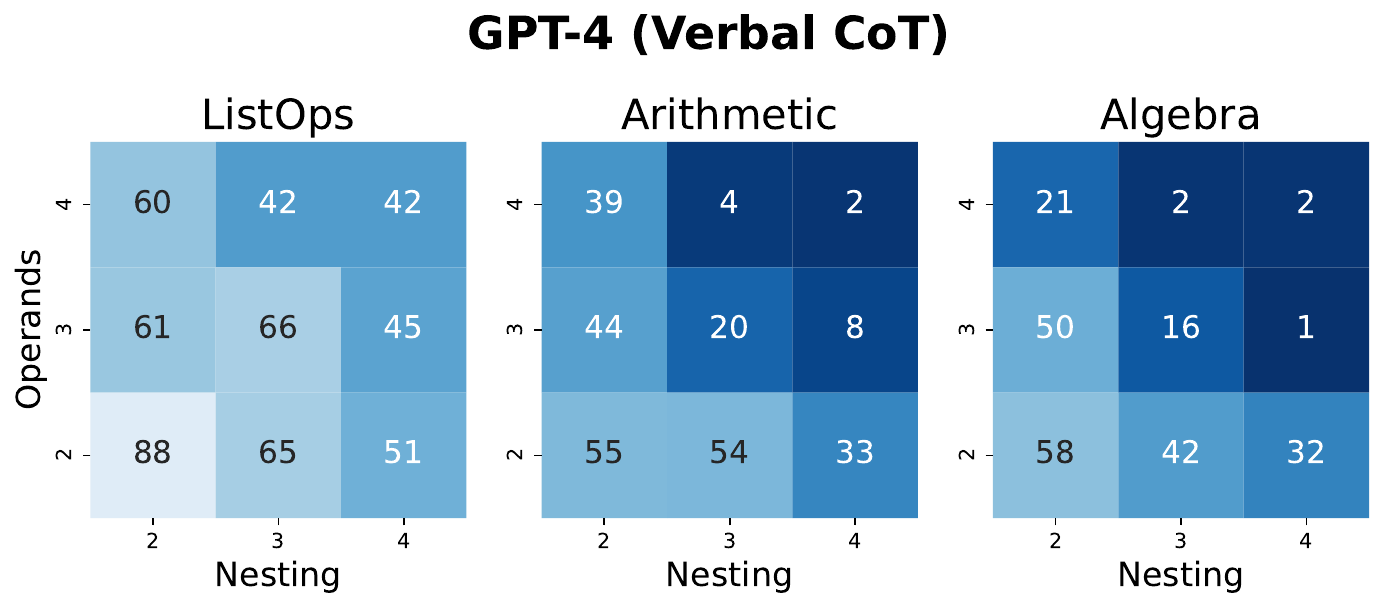}
    \end{subfigure}
    \hfill
    \begin{subfigure}[b]{0.48\textwidth}
        \centering
        \includegraphics[width=\textwidth]{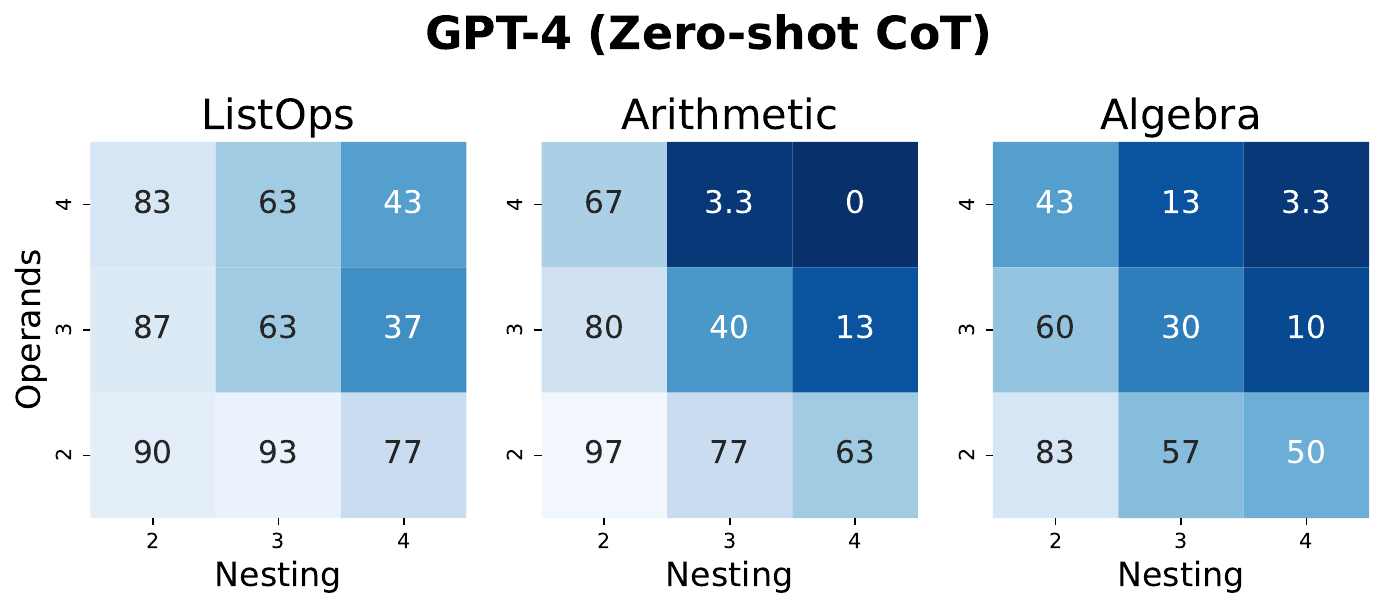}
    \end{subfigure}

    \begin{subfigure}[b]{0.48\textwidth}
        \centering
        \includegraphics[width=\textwidth]{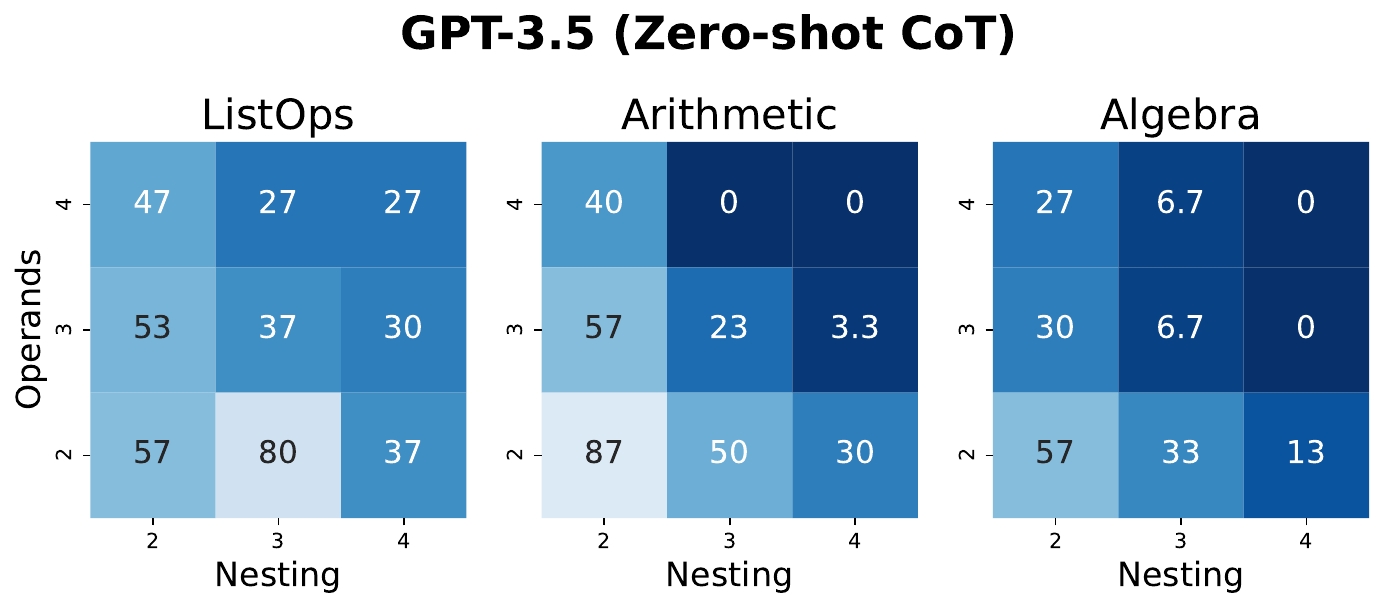}
    \end{subfigure}
\end{figure}

\newpage
\subsection{Examples of prompts}
\label{subpar:app2}
\subsubsection{ListOps}
\input{tables/appendix/prompts_table_listops}

\subsubsection{Arithmetic}
\input{tables/appendix/prompts_table_arithmetic}

\subsubsection{Algebra}
\input{tables/appendix/prompts_table_algebra}

%% file: tables/appendix/prompts_table_listops.tex
\begin{longtable}{P{2.2cm}P{11.8cm}}
Prompt type & Example \\
\hline
Zero-shot & Q: MIN, MAX and SM are operators on lists of single-digit integers which have the semantics of minimum, maximum and sum modulo 10, respectively. Solve the following expression involving these operators:
[MIN[MAX[MIN68]8][MAX[SM23]6]].

A: The final result is (arabic numeral): \\
\hline
Zero-shot role & Q: MIN, MAX and SM are operators on lists of single-digit integers which have the semantics of minimum, maximum and sum modulo 10, respectively. Solve the following expression involving these operators:
[MIN[MAX[MIN68]8][MAX[SM23]6]].

A: The final result is (arabic numeral): \\
\hline
Few-shot & Q: MIN, MAX and SM are operators on lists of single-digit integers which have the semantics of minimum, maximum and sum modulo 10, respectively. Solve the following expression involving these operators:
[MIN37].

A: [MIN37]=3.
\linebreak
\linebreak Q: MIN, MAX and SM are operators on lists of single-digit integers which have the semantics of minimum, maximum and sum modulo 10, respectively. Solve the following expression involving these operators:
[MAX[MIN41]2].

A: [MAX[MIN41]2]=2.
\linebreak
\linebreak Q: MIN, MAX and SM are operators on lists of single-digit integers which have the semantics of minimum, maximum and sum modulo 10, respectively. Solve the following expression involving these operators:
[SM[SM794][SM498]7].

A: [SM[SM794][SM498]7]=8.
\linebreak
\linebreak Q: MIN, MAX and SM are operators on lists of single-digit integers which have the semantics of minimum, maximum and sum modulo 10, respectively. Solve the following expression involving these operators:
[MIN[MAX[MIN68]8][MAX[SM23]6]]. \\
\hline
Symbolic CoT & Q: MIN, MAX and SM are operators on lists of single-digit integers which have the semantics of minimum, maximum and sum modulo 10, respectively. Solve the following expression involving these operators:
[SM73].

A: [SM73]=
0.
\linebreak
\linebreak Q: MIN, MAX and SM are operators on lists of single-digit integers which have the semantics of minimum, maximum and sum modulo 10, respectively. Solve the following expression involving these operators:
[SM[SM86]1].

A: [SM[SM86]1]=
[SM41]=
5.
\linebreak
\linebreak Q: MIN, MAX and SM are operators on lists of single-digit integers which have the semantics of minimum, maximum and sum modulo 10, respectively. Solve the following expression involving these operators:
[MIN[MAX243]4[MAX937]].

A: [MIN[MAX243]4[MAX937]]=
[MIN[MAX243]49]=
[MIN449]=
4.
\linebreak
\linebreak Q: MIN, MAX and SM are operators on lists of single-digit integers which have the semantics of minimum, maximum and sum modulo 10, respectively. Solve the following expression involving these operators:
[MAX[SM[SM85][MIN84]][SM3[MIN40]]]. \\
\hline
Verbal CoT & Q: MIN, MAX and SM are operators on lists of single-digit integers which have the semantics of minimum, maximum and sum modulo 10, respectively. Solve the following expression involving these operators:
[MIN82].

A: Let's solve the following expression: [MIN82].
Simplifying the expression, we get to the final result: 2
\linebreak
\linebreak Q: MIN, MAX and SM are operators on lists of single-digit integers which have the semantics of minimum, maximum and sum modulo 10, respectively. Solve the following expression involving these operators:
[MIN[SM56][MAX87]].

A: Let's solve the following expression: [MIN[SM56][MAX87]].
Simplifying an expression without nested parentheses, we get: [MIN[SM56]8].
Simplifying the expression, it becomes: [MIN18]
Taking an immediate solution step, we get to the final result: 1.
\linebreak
\linebreak Q: MIN, MAX and SM are operators on lists of single-digit integers which have the semantics of minimum, maximum and sum modulo 10, respectively. Solve the following expression involving these operators:
[MIN[MIN326]0[SM851]].

A: Let us recall the expression to be solved: [MIN[MIN326]0[SM851]].
By solving a simple expression, we obtain: [MIN[MIN326]04].
Solving a expression within a single pair of brackets, we get: [MIN204].
Taking an immediate solution step, we get to the final result: 0.
\linebreak
\linebreak Q: MIN, MAX and SM are operators on lists of single-digit integers which have the semantics of minimum, maximum and sum modulo 10, respectively. Solve the following expression involving these operators:
[MIN[MAX[MIN68]8][MAX[SM23]6]]. \\
\hline
Zero-shot CoT & Q: MIN, MAX and SM are operators on lists of single-digit integers which have the semantics of minimum, maximum and sum modulo 10, respectively. Solve the following expression involving these operators:
[MIN[MAX[MIN68]8][MAX[SM23]6]].

A: Let's think step-by-step. \\
\hline
\end{longtable}

%% file: tables/appendix/prompts_table_arithmetic.tex
\begin{longtable}{P{2.2cm}P{11.8cm}}
Prompt type & Example \\
\hline
Zero-shot & Q: Solve the following arithmetic expression computing the modulo 100 of each intermediate value if it's positive, and the modulo -100 if it's negative:
(-66-(-84*(-34+0))).

A: The final result is (arabic numerals): \\
\hline
Zero-shot role & Q: Solve the following arithmetic expression computing the modulo 100 of each intermediate value if it's positive, and the modulo -100 if it's negative:
(-66-(-84*(-34+0))).

A: The final result is (arabic numerals): \\
\hline
Few-shot & Q: Solve the following arithmetic expression taking each intermediate value modulo 100 if it's positive, and modulo -100 if it's negative: (51*39).

A: (51*39)=89.
\linebreak
\linebreak Q: Solve the following arithmetic expression taking each intermediate value modulo 100 if it's positive, and modulo -100 if it's negative: ((28*-53)*(-76*90)).

A: ((28*-53)*(-76*90))=60.
\linebreak
\linebreak Q: Solve the following arithmetic expression taking each intermediate value modulo 100 if it's positive, and modulo -100 if it's negative: (40-54-(-33--97+-19)).

A: (40-54-(-33--97+-19))=-59.
\linebreak
\linebreak Q: Solve the following arithmetic expression taking each intermediate value modulo 100 if it's positive, and modulo -100 if it's negative: (-66-(-84*(-34+0))). \\
\hline
Symbolic CoT & Q: Solve the following arithmetic expression taking each intermediate value modulo 100 if it's positive, and modulo -100 if it's negative: (-16*-37).

A: (-16*-37)=
92.
\linebreak
\linebreak Q: Solve the following arithmetic expression taking each intermediate value modulo 100 if it's positive, and modulo -100 if it's negative: ((87*-51)-(47*-6)).

A: ((87*-51)-(47*-6))=
((87*-51)--82)=
(-37--82)=
45.
\linebreak
\linebreak Q: Solve the following arithmetic expression taking each intermediate value modulo 100 if it's positive, and modulo -100 if it's negative: ((-12--28-74)+-21+(76+-32+-87)).

A: ((-12--28-74)+-21+(76+-32+-87))=
((-12--28-74)+-21+-43)=
(-58+-21+-43)=
-22.
\linebreak
\linebreak Q: Solve the following arithmetic expression taking each intermediate value modulo 100 if it's positive, and modulo -100 if it's negative: (((-33-39)*(67*65))-22). \\
\hline
Verbal CoT & Q: Solve the following arithmetic expression taking each intermediate value modulo 100 if it's positive, and modulo -100 if it's negative: (-14*88).

A: Let us recall the expression to be solved: (-14*88).
Simplifying the expression, we get to the final result: -32
\linebreak
\linebreak Q: Solve the following arithmetic expression taking each intermediate value modulo 100 if it's positive, and modulo -100 if it's negative: ((92*26)*(-35*59)).

A: Let us recall the expression to be solved: ((92*26)*(-35*59)).
Simplifying the expression, it becomes: ((92*26)*-65)
Solving a expression within a single pair of brackets, we get: (92*-65).
Simplifying the expression, we get to the final result: -80
\linebreak
\linebreak Q: Solve the following arithmetic expression taking each intermediate value modulo 100 if it's positive, and modulo -100 if it's negative: (83-(46+-5-54)-25).

A: We need to solve the following expression: (83-(46+-5-54)-25).
Taking an immediate solution step, we obtain: (83--13-25).
As this expression is in a simple form, we can get to the final result: 71
\linebreak
\linebreak Q: Solve the following arithmetic expression taking each intermediate value modulo 100 if it's positive, and modulo -100 if it's negative: (-66-(-84*(-34+0))). \\
\hline
Zero-shot CoT & Q: Solve the following arithmetic expression computing the modulo 100 of each intermediate value if it's positive, and the modulo -100 if it's negative:
(((-33-39)*(67*65))-22).

A: Let's think step-by-step. \\
\hline
\end{longtable}

%% file: tables/appendix/prompts_table_algebra.tex
\begin{longtable}{P{2.2cm}P{11.8cm}}
Prompt type & Example \\
\hline
Zero-shot & Q: Simplify the following algebraic expression, computing the modulo 100 of the numerical coefficient of each intermediate value if it's positive, and the modulo -100 if it's negative:
((-78*b*x*y+(+50*b*x*y+-22*b*x*y))+((-b*x+-57*b*x)+(-38*b*x+-99*b*x))). If possible, factor by grouping the final result.

A: The final result is (algebraic expression): \\
\hline
Zero-shot role & Q: Simplify the following algebraic expression, computing the modulo 100 of the numerical coefficient of each intermediate value if it's positive, and the modulo -100 if it's negative:
(-85*a*y+((-98*a*x*y+2*a*x*y)+0*x*y*a)).
If possible, factor by grouping the final result.

A: The final result is (algebraic expression): \\
\hline
Few-shot & Q: Solve the following algebraic expression taking the numerical coefficient of each intermediate value modulo 100 if it's positive, and modulo -100 if it's negative:
(-55*b*x*y+-8*b*x).
If possible, factor by grouping the final result.

A: (-55*b*x*y+-8*b*x)=-b*x*(55*y+8).
\linebreak
\linebreak Q: Solve the following algebraic expression taking the numerical coefficient of each intermediate value modulo 100 if it's positive, and modulo -100 if it's negative:
((-54*x*y+-68*x*y)+(-99*x*y++62*x*y)).
If possible, factor by grouping the final result.

A: ((-54*x*y+-68*x*y)+(-99*x*y++62*x*y))=-59*x*y.
\linebreak
\linebreak Q: Solve the following algebraic expression taking the numerical coefficient of each intermediate value modulo 100 if it's positive, and modulo -100 if it's negative:
((+12*x*y+-59*x*y++58*x*y)+(+36*x*y++13*x*y++93*x*y)+(+96*x*y+-55*x*y++73*x*y)).
If possible, factor by grouping the final result.

A: ((+12*x*y+-59*x*y++58*x*y)+(+36*x*y++13*x*y++93*x*y)+(+96*x*y+-55*x*y++73*x*y))=+67*x*y.
\linebreak
\linebreak Q: Solve the following algebraic expression taking the numerical coefficient of each intermediate value modulo 100 if it's positive, and modulo -100 if it's negative:
(-85*a*y+((-98*a*x*y+2*a*x*y)+0*x*y*a)).
If possible, factor by grouping the final result. \\
\hline
Symbolic CoT & Q: Solve the following algebraic expression taking the numerical coefficient of each intermediate value modulo 100 if it's positive, and modulo -100 if it's negative:
(+39*a*b*y++15*a*b*x*y).
If possible, factor by grouping the final result.

A: (+39*a*b*y++15*a*b*x*y)=
+3*a*b*y*(5*x+13).
\linebreak
\linebreak Q: Solve the following algebraic expression taking the numerical coefficient of each intermediate value modulo 100 if it's positive, and modulo -100 if it's negative:
(+21*x*y+(-26*x*y+-92*x*y)).
If possible, factor by grouping the final result.

A: (+21*x*y+(-26*x*y+-92*x*y))=
(+21*x*y+-18*x*y)=
+3*x*y.
\linebreak
\linebreak Q: Solve the following algebraic expression taking the numerical coefficient of each intermediate value modulo 100 if it's positive, and modulo -100 if it's negative:
((+45*b*x++22*b*x+-47*b*x)+-62*b*x*y).
If possible, factor by grouping the final result.

A: ((+45*b*x++22*b*x+-47*b*x)+-62*b*x*y)=
(+20*b*x+-62*b*x*y)=
-2*b*x*(31*y-10).
\linebreak
\linebreak Q: Solve the following algebraic expression taking the numerical coefficient of each intermediate value modulo 100 if it's positive, and modulo -100 if it's negative:
((-78*b*x*y+(+50*b*x*y+-22*b*x*y))+((-b*x+-57*b*x)+(-38*b*x+-99*b*x))).
If possible, factor by grouping the final result. \\
\hline
Verbal CoT & Q: Solve the following algebraic expression taking the numerical coefficient of each intermediate value modulo 100 if it's positive, and modulo -100 if it's negative:
(+10*a*b*x*y+-23*a*b*x).
If possible, factor by grouping the final result.

A: The expression we need to solve is: (+10*a*b*x*y+-23*a*b*x).
Simplifying the expression and factoring by grouping, we get to the final result: +a*b*x*(10*y-23)
\linebreak
\linebreak Q: Solve the following algebraic expression taking the numerical coefficient of each intermediate value modulo 100 if it's positive, and modulo -100 if it's negative:
(-8*a*x*y+(-38*a*x+-70*a*x)).
If possible, factor by grouping the final result.

A: Let us recall the expression to be solved: (-8*a*x*y+(-38*a*x+-70*a*x)).
By solving a simple expression, we obtain: (-8*a*x*y+-8*a*x).
As this expression is in a simple form, we can get to the final result factoring by grouping: -8*a*x*(y+1)
\linebreak
\linebreak Q: Solve the following algebraic expression taking the numerical coefficient of each intermediate value modulo 100 if it's positive, and modulo -100 if it's negative:
(+31*a*b*y+(-50*a*b*x+-64*a*b*x+-46*a*b*x)).
If possible, factor by grouping the final result.

A: We need to solve the following expression: (+31*a*b*y+(-50*a*b*x+-64*a*b*x+-46*a*b*x)).
Taking an immediate solution step, we obtain: (+31*a*b*y+-60*a*b*x).
Taking an immediate solution step and factoring by grouping, we get to the final result: -a*b*(60*x-31*y).
\linebreak
\linebreak Q: Solve the following algebraic expression taking the numerical coefficient of each intermediate value modulo 100 if it's positive, and modulo -100 if it's negative:
((-78*b*x*y+(+50*b*x*y+-22*b*x*y))+((-b*x+-57*b*x)+(-38*b*x+-99*b*x))).
If possible, factor by grouping the final result. \\
\hline
Zero-shot CoT & Q: Simplify the following algebraic expression, computing the modulo 100 of the numerical coefficient of each intermediate value if it's positive, and the modulo -100 if it's negative:
(-85*a*y+((-98*a*x*y+2*a*x*y)+0*x*y*a)).
If possible, factor by grouping the final result.

A: Let's think step-by-step. \\
\hline
\end{longtable}